\documentclass[letterpaper, 10 pt, conference]{ieeeconf}  

\IEEEoverridecommandlockouts                              

\usepackage{graphicx} 
\usepackage{amsmath} 
\usepackage{amssymb}  
\usepackage{bm}
\usepackage{lipsum}
\usepackage{mathtools}

\usepackage{url}

\usepackage{balance}
\usepackage{xcolor}
\usepackage{setspace}

\usepackage[linesnumbered,ruled,vlined]{algorithm2e}

\SetCommentSty{mycommfont}
\SetKwInput{KwInput}{Input}                
\SetKwInput{KwOutput}{Output}  
\SetKwInput{KwParam}{Parameter}  

\DeclareMathOperator*{\argmin}{\arg\!\min}

\title{\LARGE \bf
Robust Single Rotation Averaging Revisited
}

\usepackage[hidelinks]{hyperref}

\author{Seong Hun Lee and Javier Civera
\thanks{*This work was supported by the Spanish
Government (projects PID2021-127685NB-I00 and TED2021-131150BI00) and the Arag\'on Government (project DGA-T45 23R).}
\thanks{The authors are with I3A, University of Zaragoza, 50018, Spain
        {\tt\small seonghunlee@unizar.es;jcivera@unizar.es}}%
}

\begin{document}

\maketitle
\thispagestyle{empty}
\pagestyle{empty}

\begin{abstract}
In this work, we propose a novel method for robust single rotation averaging that can efficiently handle an extremely large fraction of outliers.
Our approach is to minimize the total truncated least unsquared deviations (TLUD) cost of geodesic distances.
The proposed algorithm consists of three steps:
First, we consider each input rotation as a potential initial solution and choose the one that yields the least sum of truncated chordal deviations.
Next, we obtain the inlier set using the initial solution and compute its chordal $L_2$-mean.
Finally, starting from this estimate, we iteratively compute the geodesic $L_1$-mean of the inliers using the Weiszfeld algorithm on $SO(3)$.
An extensive evaluation shows that our method is robust against up to 99\% outliers given a sufficient number of accurate inliers, outperforming the current state of the art.
\end{abstract}

\section{INTRODUCTION}
\begin{figure*}[t]
 \centering
 \includegraphics[width=0.8\textwidth]{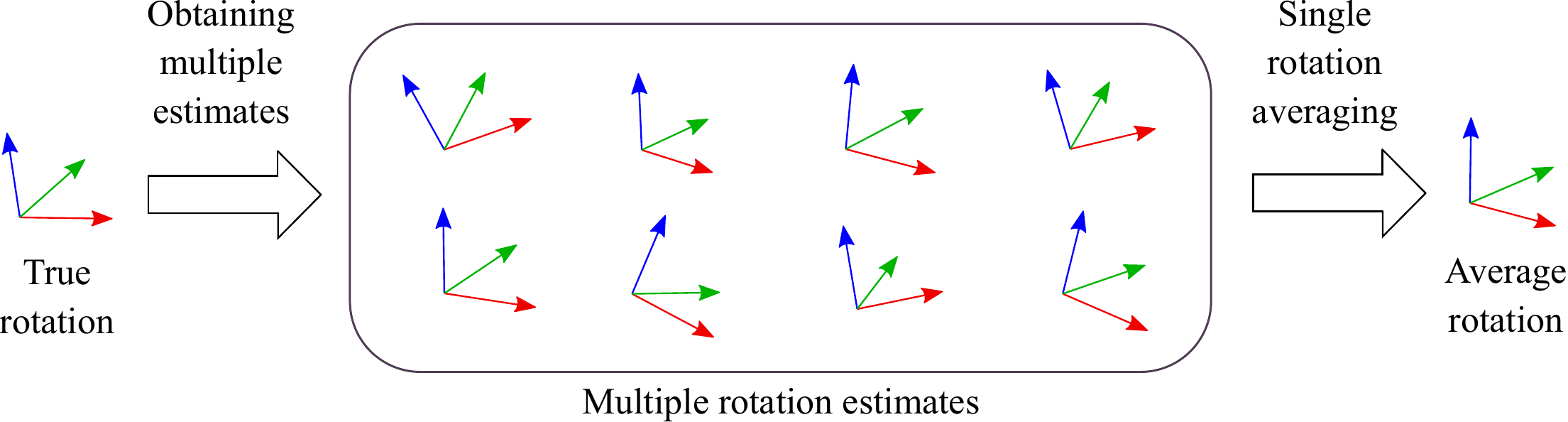}
\caption{Synopsis of the single rotation averaging problem.
The unknown rotation can be estimated by averaging multiple noisy (and possibly outlier-contaminated) estimates of it.}
\label{fig:sra}
\end{figure*}

We consider the problem of robust single rotation averaging, \textit{i.e.}, averaging several noisy and outlier-contaminated estimates of a single rotation to obtain the most accurate estimate (see Fig. \ref{fig:sra}).
This problem is relevant in a wide variety of other problems and applications, such as vision-based robot localization \cite{maggio_2023_icra}, simultaneous localization and mapping (SLAM) \cite{joo_2020_icra}, object pose estimation \cite{yang_2023_cvpr}, multiple rotation averaging and structure from motion (SfM) \cite{hartley2011L1, lee_2022_cvpr}, non-rigid SfM \cite{kumar_2022_eccv}, point cloud registration \cite{sun_2022_access}, camera rig calibration \cite{dai2009rotation}, motion capture \cite{inna2010arithmetic}, spacecraft attitude determination \cite{lam2007precision,markley2007averaging} and crystallography \cite{humbert1996determination, morawiec1998note}. 

Typically, single rotation averaging involves the minimization of a cost function based on the distances to the input rotations.
We refer to \cite{hartley2013rotation} for an excellent study of the several different distance functions commonly used in literature.
In the presence of outliers, a robust cost function can be used to reduce their adverse impact on the estimation accuracy.
For instance, using the $L_1$ or Huber-like cost function is shown to be more robust than using the $L_2$ function \cite{hartley2011L1,aftab2015convergence}.
However, these functions are not suitable when the outlier ratio is extremely high (\textit{e.g.}, over 90\%) \cite{lajoie_2019_ral}.
To handle a large fraction of outliers, one could use a truncated cost function which assigns zero weights to residuals above a certain threshold.
Examples include Tukey's biweight function \cite{tukey} and a truncated least squares (TLS) cost \cite{yang_2020_ral}.

The main difficulty that arises with iterative algorithms using such cost functions is that the initial guess must be sufficiently accurate in order for the solution to converge to the global optimum.
At high outlier ratios, an accurate initialization can be a challenge.

\begin{figure}[t]
 \centering
 \includegraphics[width=0.45\textwidth]{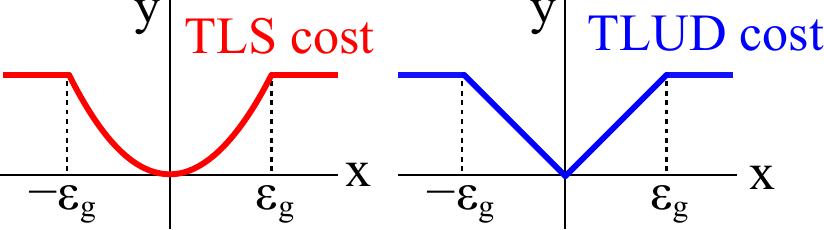}
\caption{The TLS \cite{yang_2020_ral} and the TLUD cost \eqref{eq:TLUD}. In this work, we formulate the single rotation averaging problem as minimization of the total TLUD cost under the geodesic metric.}
\label{fig:TLUD}
\end{figure}

In this work, we propose a novel single rotation averaging method that is robust to an extremely large proportion of outliers (up to 99\%).
Our method aims to minimize the total truncated least unsquared deviations (TLUD) cost of geodesic distances (see Fig. \ref{fig:TLUD}).
The novel contribution of this work is a simple yet effective initialization of the solution: 
Try each and every input rotation to find the one producing the lowest \textit{proxy} cost.
That is, we minimize the truncated sum of chordal deviations (the proxy cost), instead of the truncated sum of geodesic deviations (the actual cost we want to minimize use in the end).
Once the initial solution and corresponding inlier set is found, we obtain the geodesic $L_1$-mean using the Weiszfeld algorithm proposed in \cite{hartley2011L1}.
We conduct an extensive evaluation on both synthetic and real data, demonstrating that our method leads to state-of-the-art results in single rotation averaging.
Our code is publicly available at 
{\color{magenta}\url{https://github.com/sunghoon031/SingleRotationAveraging_TLUD}}.


\section{RELATED WORK}
\label{sec:related}
In the absence of outliers, the geodesic or chordal $L_2$-mean can be a reasonable choice for averaging rotations.
Manton \cite{manton_2004_icarcv} proposes a convergent algorithm for computing the geodesic $L_2$-mean.
On the other hand, the chordal $L_2$-mean can be found in closed form using either the quaternion representation \cite{markely_2007_jgcd} or matrix representation of rotations \cite{sarlette_2009_siam}.
The problem with the $L_2$-mean is that it is highly sensitive to outliers.
In terms of robustness, it is well understood that the $L_1$-mean (\textit{i.e.}, median) is generally a better choice than the $L_2$-mean.
Hartley et al. \cite{hartley2011L1} adapt the Weiszfeld algorithm iteratively in tangent spaces of $SO(3)$ to obtain the geodesic $L_1$-mean of rotations.
Their solution is shown to be significantly better than the $L_2$-mean.

More recently, Lee and Civera \cite{robust_single_rotation_averaging} propose an outlier rejection scheme that runs at each iteration of the standard Weiszfeld algorithm \cite{weiszfeld1,weiszfeld2} to achieve higher robustness than the geodesic $L_1$-mean.
For efficiency, they first embed the rotations in a Euclidean space $\mathbb{R}^9$, compute the average, and project it onto $SO(3)$ at the end.
Consequently, they achieve a more robust and efficient estimation than Hartley's method \cite{hartley2011L1}.
Yang and Carlone \cite{yang_2020_nips} propose a certifiably robust algorithm that minimizes the TLS cost. 
Using their method, they show that they can handle up to 80\% outliers in single rotation averaging.
Shi et al. \cite{shi_2021_icra} propose a graph-theoretic approach to reject outliers by checking the compatibility between pairs of rotations.
Then, by finding the maximum clique or maximum $k$-core of the graph, they can prune a large fraction of outliers and achieve even greater robustness than \cite{robust_single_rotation_averaging}.
Mankovich and Birdal \cite{mankovich_2023_arxiv} propose an IRLS algorithm for computing the flag-median under the chordal metric.
By applying their algorithm for single rotation averaging, they show that their method is robust against 80\% outliers.

\section{PRELIMINARIES}
\label{sec:prelim}
Given a 3D vector $\mathbf{v}$, we define $\mathbf{v}^\wedge$ as the corresponding $3\times3$ skew-symmetric matrix, and denote the inverse operator by $(\cdot)^\vee$, i.e., $\left(\mathbf{v}^\wedge\right)^\vee=\mathbf{v}$.
The Euclidean and the Frobenius norm are denoted by $\lVert \cdot \rVert$ and $\lVert \cdot \rVert_F$, respectively.
A rotation can be represented by a rotation matrix $\mathbf{R}\in SO(3)$ or a rotation vector $\mathbf{v}=\theta\hat{\mathbf{v}}$ where $\theta$ and $\hat{\mathbf{v}}$ represent the angle and the unit axis of the rotation, respectively.
The two representations are related by Rodrigues formula, and we denote the corresponding mapping between them by $\text{Exp}(\cdot)$ and $\text{Log}(\cdot)$ \cite{forster2017onmanifold}:
\begin{equation}
    \mathbf{R}=\mathrm{Exp}(\mathbf{v})
    :=
    \mathbf{I}+\frac{\sin\left(\lVert\mathbf{v}\rVert\right)}{\lVert\mathbf{v}\rVert}\mathbf{v}^\wedge+\frac{1-\cos\left(\lVert\mathbf{v}\rVert\right)}{\lVert\mathbf{v}\rVert^2}\left(\mathbf{v}^\wedge\right)^2,
\end{equation}
\begin{equation}
\label{eq:log_map1}
    \mathbf{v}=\mathrm{Log}(\mathbf{R}):=
    \frac{\theta}{2\sin(\theta)}\left(\mathbf{R}-\mathbf{R}^\top\right)^\vee 
\end{equation}
\begin{equation}
\label{eq:log_map2}
    \text{with} \quad \theta = \cos^{-1}\left(\frac{\mathrm{tr}(\mathbf{R})-1  }{2}\right).
\end{equation}
The geodesic distance between two rotations $\mathbf{R}_1$ and $\mathbf{R}_2$, \textit{i.e.}, $d_\mathrm{g}(\mathbf{R}_1, \mathbf{R}_2)$, is obtained by substituting $\mathbf{R}_1\mathbf{R}_2^\top$ into $\mathbf{R}$ in \eqref{eq:log_map2}.
The chordal and the geodesic distance are related by the following equation \cite{hartley2013rotation}:
\begin{align}
    d_\mathrm{c}(\mathbf{R}_1, \mathbf{R}_2):=& \lVert\mathbf{R}_1-\mathbf{R}_2\rVert_F\\
    =&2\sqrt{2}\sin\left(\frac{d_\mathrm{g}(\mathbf{R}_1, \mathbf{R}_2)}{2}\right) \label{eq:chordal}.
\end{align}
We define $\mathrm{proj}_{SO(3)}(\cdot)$ as the projection of the $3\times3$ matrix onto the special orthogonal group $SO(3)$, which gives the closest rotation in the Frobenius norm \cite{arun}.
For $\mathbf{M}\in\mathbb{R}^{3\times3}$,
\begin{equation}
\label{eq:orthogonal_projection}
    \mathrm{proj}_{SO(3)}(\mathbf{M}):=\mathbf{UWV}^\top,
\end{equation}
where
\begin{gather}
    \mathbf{U}\bm{\Sigma}\mathbf{V}^\top 
    = \mathrm{SVD}\left(\mathbf{M}\right),  \\
    \mathbf{W}
    = 
    \mathrm{diag}\left(1, 1, \mathrm{sign}\left(\det\left(\mathbf{UV}^\top\right)\right)\right).
\end{gather}

\section{METHOD}
\label{sec:method}
In order to handle a large fraction of outliers, we formulate the rotation averaging problem as follows:
\begin{equation}
\label{eq:problem}
    \mathbf{R}_\mathrm{avg} = \argmin_\mathbf{R} \sum_{i=1}^N \rho\left(d_\mathrm{g}(\mathbf{R}_i, \mathbf{R}), \epsilon_\mathrm{g}\right),
\end{equation}
where $\rho(\cdot, \cdot)$ is a TLUD cost, \textit{i.e.},
\begin{equation}
\label{eq:TLUD}
    \rho(x, \epsilon_g) = \begin{cases}
        |x| \quad \text{if} \ |x| \leq \epsilon_\mathrm{g}\\
        \epsilon_\mathrm{g} \quad \text{otherwise}
    \end{cases}
\end{equation}
where $\epsilon_g$ is the inlier threshold of the geodesic distance.
This function is drawn in Fig. \ref{fig:TLUD}.
Drawn in the same figure is a TLS cost, another commonly used function in robust optimization. 
The difference is that the inlier regions of the TLS and the TLUD cost are represented by a quadratic and a linear function, respectively.
We chose the TLUD cost over the TLS cost because on average the $L_2$-mean is known to be less accurate than the $L_1$-mean in outlier-free cases of single rotation averaging \cite{robust_single_rotation_averaging}.

The problem \eqref{eq:problem} can be solved using the Weiszfeld algorithm on $SO(3)$ proposed in \cite{hartley2011L1}, provided that the inlier set $\mathcal{I}$ is known, \textit{i.e.,} 
\begin{equation}
\label{eq:inlier_set1}
    \mathcal{I} = \{i \ | \ d_\mathrm{g}(\mathbf{R}_i,\mathbf{R}_\mathrm{avg}) \leq \epsilon_\mathrm{g}, i = 1, \cdots, N \}.
\end{equation}
Unfortunately, the problem becomes circular, as the acquisition of $\mathcal{I}$ requires the knowledge of $\mathbf{R}_\mathrm{avg}$, and vice versa.
We can circumvent this deadlock by making a good initial guess of $\mathbf{R}_\mathrm{avg}$ and then finding the corresponding initial inlier set.
Then, we can alternate between the computation of the inlier set \eqref{eq:inlier_set1} and $\mathbf{R}_\mathrm{avg}$ until the solution converges.
In practice, however, we found that this alternation is not necessary if the initial guess of $\mathbf{R}_\mathrm{avg}$ is reasonably accurate.
Hence, we focus on the accurate initialization of $\mathbf{R}_\mathrm{avg}$ in the remainder of this section.

\begin{algorithm}[t]
\setstretch{1.1}
\caption{Proposed Single Rotation Averaging}
\label{al:ours}
\DontPrintSemicolon
  \KwInput{List of rotation matrices $\{\mathbf{R}_i\}_{i=1}^N$.}
  \KwOutput{$\mathbf{R}_\mathrm{avg}$.}
  \KwParam{Inlier threshold $\epsilon_\mathrm{c}$, Update threshold $\delta$, Maximum number of iterations $\mathrm{it}_\mathrm{max}$. By default, we set $(\epsilon_\mathrm{c}, \delta, \mathrm{it}_\mathrm{max}) = (0.5, 0.001, 10)$.}
  \tcc{Find the inlier set:}

  $\displaystyle \mathbf{R}_\mathrm{init} \gets  \argmin_{\displaystyle\mathbf{R}_{j\in\{1 \cdots N\}}}\sum_{i=1}^N \min\left(\epsilon_\mathrm{c}, \lVert\mathbf{R}_i-\mathbf{R}_j\rVert_F \right)$;

  $\mathcal{I} \gets \{i \ | \ \lVert\mathbf{R}_i-\mathbf{R}_\mathrm{init}\rVert_F < \epsilon_\mathrm{c}, i = 1, \cdots, N \}$;

 \tcc{Set the chordal $L_2$-mean \cite{hartley2013rotation} of inliers as the initial estimate:}
 
 $\mathbf{R}_\mathrm{avg} \gets \mathrm{proj}_{SO(3)}\left(\sum_{i\in\mathcal{I}}\mathbf{R}_i\right)$;
  
  \tcc{Run the Weiszfeld algorithm on SO(3) \cite{hartley2011L1} over the inlier set:}
    
  \For{$\mathrm{it}=1,2,3, \cdots, \mathrm{it}_\mathrm{max}$}
  { 
    $\mathbf{v}_i \gets \mathrm{Log}\left(\mathbf{R}_i\mathbf{R}_\mathrm{avg}^\top\right) \ \text{for} \  i\in\mathcal{I}$; 
    
    $\displaystyle \Delta\mathbf{v} \gets\frac{\sum_{i\in\mathcal{I}}\mathbf{v}_i/\lVert\mathbf{v}_i\rVert }{\sum_{i\in\mathcal{I}} 1/\lVert\mathbf{v}_i\rVert }$;
    
    $\mathbf{R}_\mathrm{avg}\gets\mathrm{Exp}(\Delta\mathbf{v})\mathbf{R}_\mathrm{avg}$;
    
    \If{$\lVert\Delta\mathbf{v}\rVert< \delta$}
    {
        Break;
    }
  }
  \Return{$\mathbf{R}_\mathrm{avg}$}
\end{algorithm}

The key idea behind our initialization method is very simple:
We try each and every input rotation and find the one that produces the smallest cost, \textit{i.e.},
\begin{equation}
\label{eq:R_init1}
    \mathbf{R}_\mathrm{init} =\argmin_{\displaystyle\mathbf{R}_{j\in\{1, \cdots, N\}}}\sum_{i=1}^N \rho\left(d_\mathrm{g}(\mathbf{R}_i, \mathbf{R}_j), \epsilon_g\right).
\end{equation}
For efficiency, we approximate the solution to \eqref{eq:R_init1} by the following rotation:
\begin{equation}
\label{eq:R_init2}
    \mathbf{R}_\mathrm{init} \approx \argmin_{\displaystyle\mathbf{R}_{j\in\{1, \cdots, N\}}}\sum_{i=1}^N \rho\left(d_\mathrm{c}(\mathbf{R}_i, \mathbf{R}_j), \epsilon_\mathrm{c}\right), 
\end{equation}
where
\begin{equation}
\label{eq:epsilons}
\epsilon_\mathrm{c}\stackrel{\eqref{eq:chordal}}{=}2\sqrt{2}\sin\left(\frac{\epsilon_\mathrm{g}}{2}\right).
\end{equation}
We call the cost function in \eqref{eq:R_init2} the \textit{proxy} cost.
To obtain our initial solution $\mathbf{R}_\mathrm{init}$, we use this proxy cost instead of the actual cost we want to minimize in the end (\textit{i.e.}, \eqref{eq:problem}).
The advantage of this approach is that we can make use of efficient matrix operations, as the computation of the chordal distance only involves arithmetic operations and square rooting.
This leads to a significant speed-up as a result:
In our setting (as described in Section \ref{sec:results}), solving \eqref{eq:R_init1} takes around $3$ seconds when $N=1000$, while solving \eqref{eq:R_init2} takes only $0.03$ seconds.

Once the initial solution is obtained by solving \eqref{eq:R_init2}, we find the inlier set as follows: 
\begin{equation}
\label{eq:inlier_set2}
    \mathcal{I} = \{i \ | \ d_\mathrm{c}(\mathbf{R}_i,\mathbf{R}_\mathrm{init}) \leq \epsilon_\mathrm{c}, i = 1, \cdots, N \}.
\end{equation}
Then, the final solution is obtained by solving the following problem using the Weiszfeld algorithm proposed in \cite{hartley2011L1}:
\begin{equation}
\label{eq:problem2}
    \mathbf{R}_\mathrm{avg} = \argmin_\mathbf{R} \sum_{i\in\mathcal{I}} d_\mathrm{g}(\mathbf{R}_i, \mathbf{R}),
\end{equation}
We use the chordal $L_2$-mean \cite{hartley2013rotation} of the inliers as the initial seed. 
Our method is summarized in Algorithm \ref{al:ours}.
In all our experiments, we set $\epsilon_c=0.5$, which corresponds to $\epsilon_g\approx 0.36$ rad (or $20.4^\circ$) according to \eqref{eq:epsilons}. 

\begin{figure*}[t]
 \centering
 \includegraphics[width=0.95\textwidth]{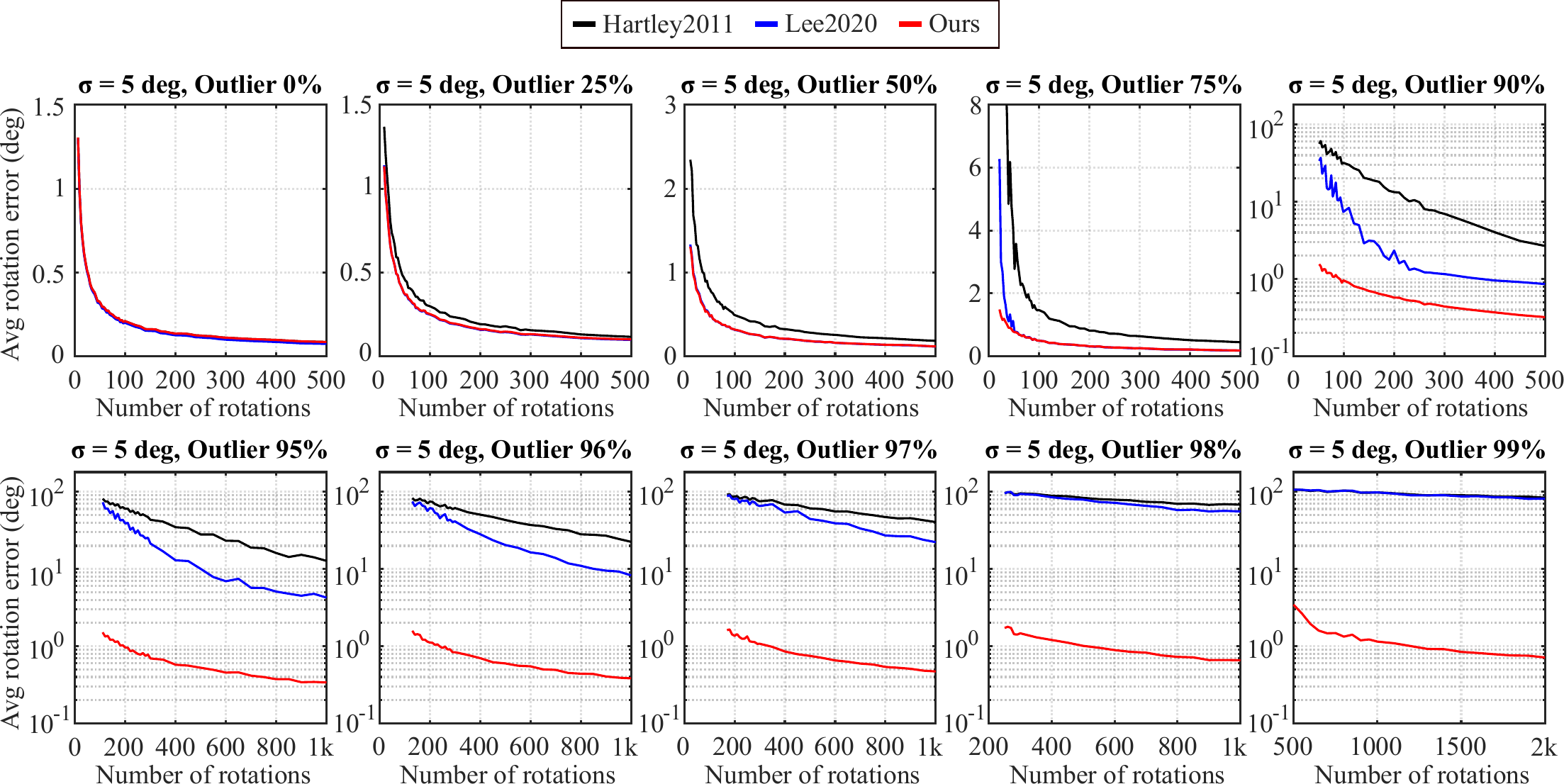}
\caption{$\bm{[\sigma=5^\circ]}$ Comparison of the average rotation errors when the inlier noise level of the simulated input rotations is 5 deg (1000 Monte Carlo runs).
The accuracy of our method is comparable to that of Lee2020 \cite{robust_single_rotation_averaging} when the outlier ratio is low. However, our method is much more robust when the outlier ratio is high.}
\label{fig:results_5deg}
\end{figure*}

\begin{figure*}[t]
 \centering
 \includegraphics[width=0.95\textwidth]{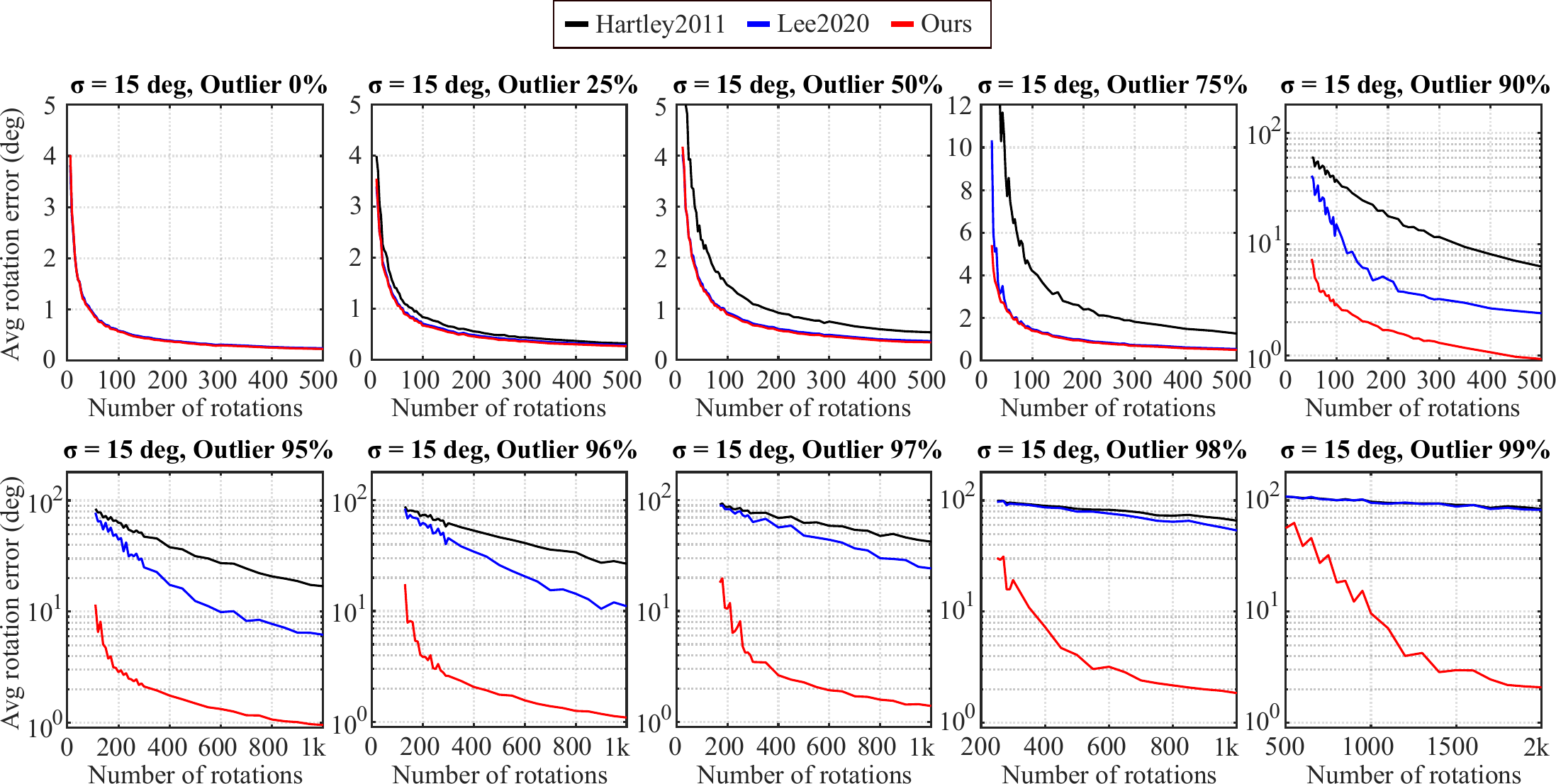}
\caption{$\bm{[\sigma=15^\circ]}$ Comparison of the average rotation errors when the inlier noise level of the simulated input rotations is 15 deg (1000 Monte Carlo runs).
The accuracy of our method is comparable to that of Lee2020 \cite{robust_single_rotation_averaging} when the outlier ratio is low. However, our method is much more robust when the outlier ratio is high.}
\label{fig:results_15deg}
\end{figure*}

\begin{figure*}[t]
 \centering
 \includegraphics[width=0.75\textwidth]{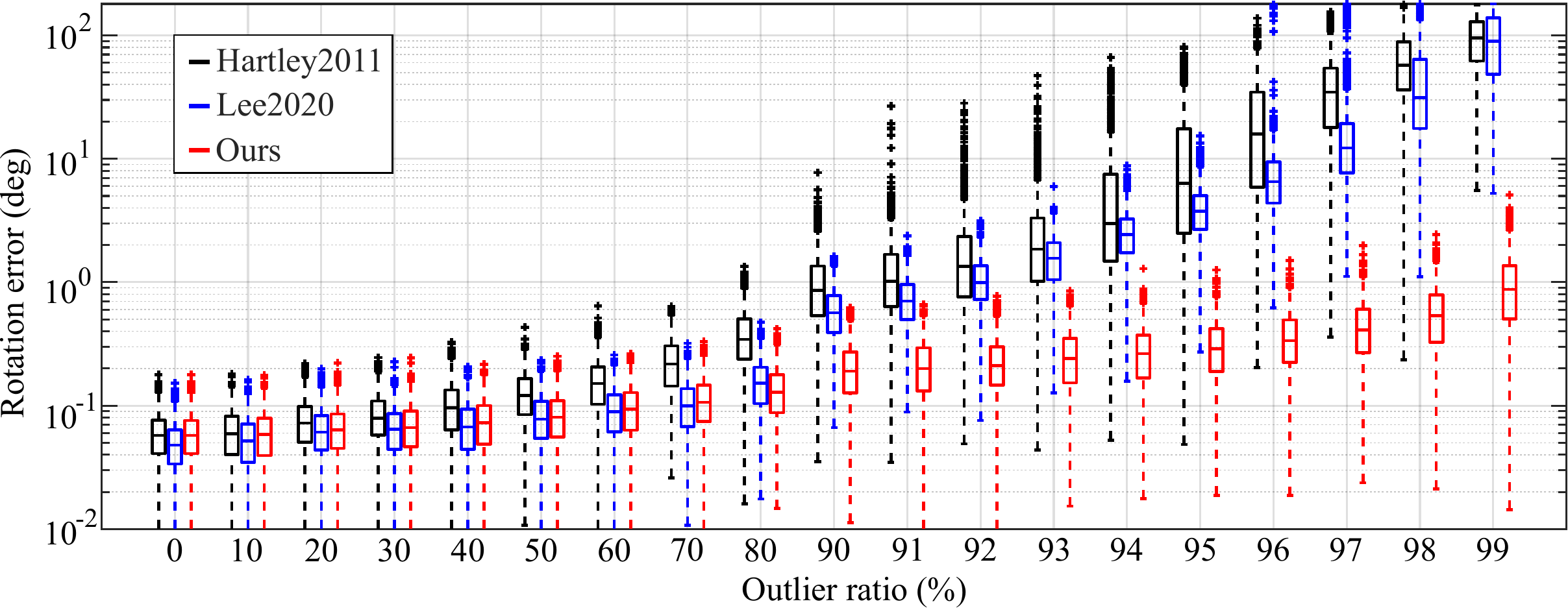}
\caption{Rotation errors when averaging 1000 rotations with inlier noise level of 5 deg (1000 Monte-Carlo runs).
Among the three methods, only ours can handle up to 99\% outliers.}
\label{fig:results_outliers}
\end{figure*}

\section{RESULTS}
\label{sec:results}
We compare our method against three existing methods: Hartley2011 \cite{hartley2011L1}, Lee2020 \cite{robust_single_rotation_averaging} and ROBIN \cite{shi_2021_icra}.
All methods are implemented in Matlab and run on a laptop with i7-4810MQ CPU (2.8 GHz) and 16 GB RAM.

{\renewcommand{\arraystretch}{1.3}%
\begin{table}[t]
\small
\begin{center}
\begin{tabular}{llll}
\hline
 & Hartley2011 \cite{hartley2011L1}& Lee2020 \cite{robust_single_rotation_averaging} & Ours    \\
\hline
\multicolumn{1}{l}{($10^2$, $0\%$)}& 0.6 &  \textbf{0.3} & 1.4 \\
\multicolumn{1}{l}{($10^2$, $50\%$)} & 0.7 & \textbf{0.4} & 1.2 \\
\multicolumn{1}{l}{($10^2$, $90\%$)} & 0.9 & \textbf{0.7} & 1.0 \\
\hline
\multicolumn{1}{l}{($10^3$, $0\%$)}& 3.0 &  \textbf{1.1} & 34 \\
\multicolumn{1}{l}{($10^3$, $50\%$)} & 3.8 & \textbf{2.2} & 39 \\
\multicolumn{1}{l}{($10^3$, $99\%$)} & \textbf{5.8} & 6.3 & 40 \\
\hline
\multicolumn{1}{l}{($10^4$, $0\%$)}& 63 &  \textbf{26} & 4128 \\
\multicolumn{1}{l}{($10^4$, $50\%$)} & 75 & \textbf{52} & 6266 \\
\multicolumn{1}{l}{($10^4$, $99.9\%$)} & \textbf{101} & 139 & 4930 \\
\hline
\end{tabular}
\end{center}
\caption{Median computation time (ms) for different numbers of rotations and outlier ratios (100 Monte-Carlo runs).
The inlier noise level is kept at 5 deg. The fastest results are highlighted in bold.}
\label{tab:timings}
\end{table}
}

\subsection{Synthetic Experiment: Ours vs. Hartley2011 vs. Lee2020}
\label{subsec:ours_vs_old}
We generate a synthetic dataset where the inlier rotations follow a Gaussian distribution with $\sigma\in\{5^\circ, 15^\circ\}$.
The outlier rotations are generated as follows:
We generate a 3D unit vector with a random direction and set it as the first column of the rotation.
Then, we generate another 3D unit vector perpendicular to the first one and set it as the second column of the rotation.
Finally, we generate the 3D unit vector by computing the cross product of the first two vectors and set it as the third column of the rotation.
While this method does not produce evenly distributed rotation angles (as in \cite{robust_single_rotation_averaging}), it prevents the concentration of outliers near the identity matrix.
The evaluation results are shown in Fig. \ref{fig:results_5deg}, \ref{fig:results_15deg} and \ref{fig:results_outliers}.
It can be seen that our method can handle up to 99\% outliers given a sufficient number of inliers.
We report the computation times in Tab. \ref{tab:timings}. 

\begin{figure}[t]
 \centering
 \includegraphics[width=0.485\textwidth]{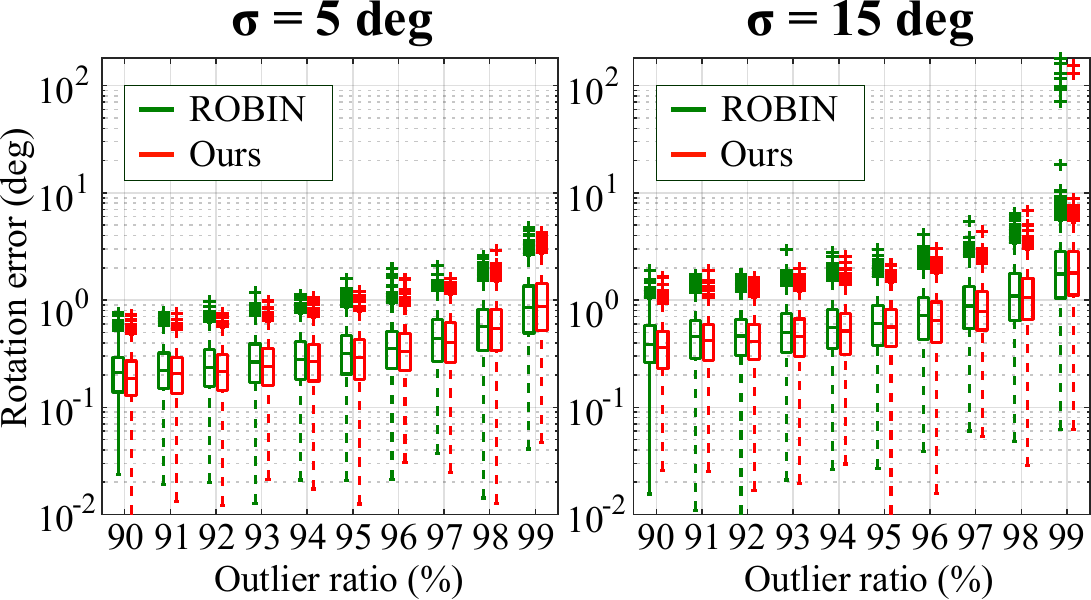}
\caption{Comparison between ROBIN \cite{shi_2021_icra} and ours when averaging 1000 rotations with two different inlier noise levels (1000 Monte-Carlo runs).
The two methods perform almost equally well.
However, at 99\% outlier ratio when $\sigma=15^\circ$, ROBIN yields large errors ($>10^\circ$) in 11 out of 1000 runs, while ours does it only twice.}
\label{fig:results_robin}
\end{figure}

\subsection{Synthetic Experiment: Ours vs. ROBIN}
\label{subsec:ours_vs_robin}
Using the same experimental set-up as in the previous section, we now compare our method with the state-of-the-art method, ROBIN \cite{shi_2021_icra}.
Note that the original implementation of ROBIN uses parallel programming in C++ for finding the maximum clique (or $k$-core) of the compatibility graph.
For a fair comparison, we implement ROBIN in Matlab using the method described in \cite{eppstein_2010_ac} for computing the maximum clique.
The same parameter is used as in the original C++ implementation.
Once the inlier set is found, we compute its geodesic $L_1$-mean using Hartley2011.

We found that ROBIN and our method have similar overall accuracy, as shown in Fig. \ref{fig:results_robin}. 
That said, our method fails much fewer times than ROBIN at 99\% outlier ratio under a large noise level.
In the original paper \cite{shi_2021_icra}, it was reported that at this outlier ratio ROBIN fails fewer times when using the maximum clique instead of the maximum $k$-core.
This suggests that our method would still be more robust than ROBIN even if the maximum $k$-core were to be used.

We also found that our method is significantly faster: Averaging 1000 rotation at 99\% outliers took 0.04s with ours and 1.82s with ROBIN (1.4s for building a compatibility graph and 0.4s finding the maximum clique).
There are two reasons for this speed difference:
First, ROBIN constructs the compatibility graph based on the geodesic distance, whereas we use the chordal distance as the proxy metric.
This enables much faster computation, as discussed earlier in Section \ref{sec:method}.
Also, ROBIN involves the additional step of maximum clique (or $k$-core) estimation after the pairwise comparison of input rotations, while we compute the cost using only additions immediately after the pairwise comparison. 
Even though computing the maximum $k$-core may take less time than computing the maximum clique, it would still incur a non-negligible additional time compared to our method that involves no such step.

\begin{figure}[t]
 \centering
 \includegraphics[width=0.4\textwidth]{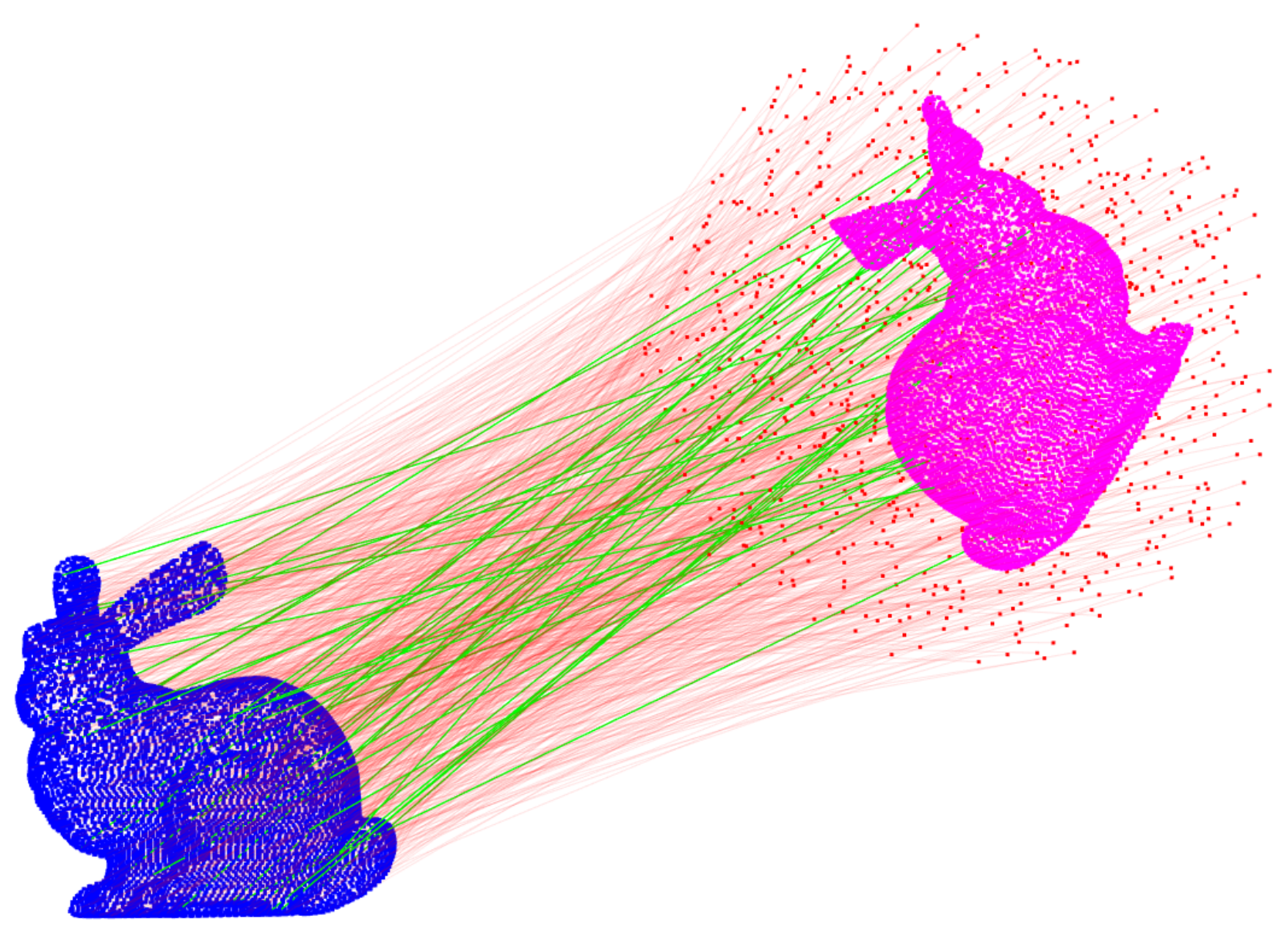}
\caption{The Bunny dataset \cite{bunny} used for the evaluation of the single rotation averaging algorithms in the context of unknown-scale point cloud registration.
The inlier correspondences are shown as the green lines and the outliers are shown as the red lines.
Here, the outlier ratio is 96\%.}
\label{fig:bunny}
\end{figure}

\subsection{Application to Point Cloud Registration}
We test the four rotation averaging methods for solving the rotation in point cloud registration.
We use the Bunny dataset from the Stanford 3D scanning repository \cite{bunny}.
We process the dataset as follows:
First, we obtain the ground-truth point cloud by downsampling the data to 1000 points and resizing it to fit inside the $[-0.5, 0.5]^3$ cube.
Next, we obtain the estimated point cloud by transforming the ground truth using a random similarity transformation with $1<s<5$, $\mathbf{R}\in SO(3)$ and $\mathbf{t}\in \mathbb{R}^3$.
Then, we add Gaussian noise of $\sigma=0.01$ and replace 0--98\% of the points with random points inside a 3D sphere of diameter $\sqrt{3}s$ (see Fig. \ref{fig:bunny}).

\begin{figure}[t]
 \centering
 \includegraphics[width=0.45\textwidth]{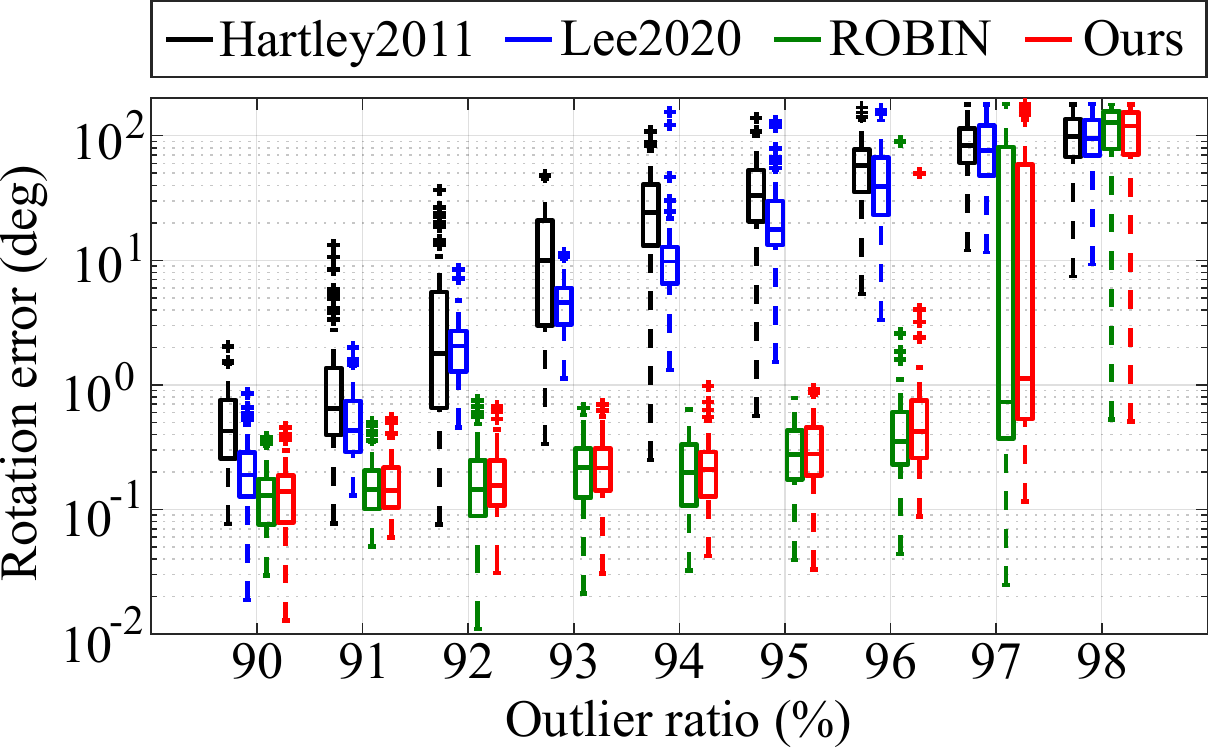}
\caption{Rotation errors when averaging 2000 rotations obtained by aligning 3-point samples from the Bunny dataset (100 Monte-Carlo runs).
Only ROBIN and our method can handle outliers up to 96\%.}
\label{fig:results_bunny}
\end{figure}

To obtain multiple estimates of the rotation between the two point clouds, we take the following approach:
\begin{enumerate}
    \item We randomly choose a set of 3 point correspondences.
    \item If they are inliers, the triangle formed by the 3 points in one point cloud should be almost similar to the corresponding triangle in the other point cloud.
    We check this by comparing the ratio of each side of the triangle.
    If the 3 ratios are similar to each other, we proceed to the next step.
    Otherwise, we discard the sample and go back to the previous step.
    \item We align the 3 points using \cite{arun} and obtain the rotation.
    \item We repeat the previous steps until we have 2000 rotation estimates.
    Afterwards, we average them using the rotation averaging algorithms.
\end{enumerate}
The results are shown in Fig. \ref{fig:results_bunny}.
Here, we again see that ROBIN and our method are more robust than the other two methods.
We emphasize, however, that we are not claiming superiority over state-of-the-art point cloud registration methods.
Instead, we are simply using this application to demonstrate that (1) single rotation averaging can be a useful tool in robust point cloud registration (as also shown in \cite{sun_2022_access}), and (2) our method is highly effective for this task.

\section{CONCLUSIONS}
\label{sec:conclusion}
In this work, we proposed a novel method for robust single rotation averaging that is capable of handling extremely large outlier ratios.
We first initialize the solution by choosing one of the input rotations that leads to the least sum of truncated chordal deviations.
Afterwards, we compute the geodesic $L_1$-mean of the inliers using the Weiszfeld algorithm \cite{hartley2011L1}.
An extensive evaluation showed that our method outperforms the state of the art and is robust against up to 99\% outliers given a sufficient number of accurate inliers.
Furthermore, we demonstrated that our method can be used to solve the rotation part of point cloud registration in the presence of a large number of outliers.

\begin{thebibliography}{10}
\providecommand{\url}[1]{#1}
\csname url@rmstyle\endcsname
\providecommand{\newblock}{\relax}
\providecommand{\bibinfo}[2]{#2}
\providecommand\BIBentrySTDinterwordspacing{\spaceskip=0pt\relax}
\providecommand\BIBentryALTinterwordstretchfactor{4}
\providecommand\BIBentryALTinterwordspacing{\spaceskip=\fontdimen2\font plus
\BIBentryALTinterwordstretchfactor\fontdimen3\font minus \fontdimen4\font\relax}
\providecommand\BIBforeignlanguage[2]{{%
\expandafter\ifx\csname l@#1\endcsname\relax
\typeout{** WARNING: IEEEtran.bst: No hyphenation pattern has been}%
\typeout{** loaded for the language `#1'. Using the pattern for}%
\typeout{** the default language instead.}%
\else
\language=\csname l@#1\endcsname
\fi
#2}}

\bibitem{maggio_2023_icra}
D.~Maggio, M.~Abate, J.~Shi, C.~Mario, and L.~Carlone, ``{Loc-NeRF}: {M}onte {C}arlo localization using neural radiance fields,'' in \emph{IEEE Intl. Conf. on Robotics and Automation (ICRA)}, 2023, pp. 4018--4025.

\bibitem{joo_2020_icra}
K.~Joo, T.-H. Oh, F.~Rameau, J.-C. Bazin, and I.~S. Kweon, ``Linear {RGB-D SLAM} for {A}tlanta world,'' in \emph{IEEE Intl. Conf. on Robotics and Automation (ICRA)}, 2020, pp. 1077--1083.

\bibitem{yang_2023_cvpr}
H.~Yang and M.~Pavone, ``Object pose estimation with statistical guarantees: Conformal keypoint detection and geometric uncertainty propagation,'' in \emph{Proceedings of the IEEE/CVF Conference on Computer Vision and Pattern Recognition (CVPR)}, 2023, pp. 8947--8958.

\bibitem{hartley2011L1}
R.~I. Hartley, K.~Aftab, and J.~Trumpf, ``{L1} rotation averaging using the {W}eiszfeld algorithm,'' in \emph{IEEE Conf. on Computer Vision and Pattern Recognition (CVPR)}, 2011, pp. 3041--3048.

\bibitem{lee_2022_cvpr}
S.~H. Lee and J.~Civera, ``{HARA}: A hierarchical approach for robust rotation averaging,'' in \emph{IEEE Conf. Comput. Vis. Pattern Recog.}, 2022, pp. 15\,777--15\,786.

\bibitem{kumar_2022_eccv}
S.~Kumar and L.~Van~Gool, ``Organic priors in non-rigid structure from motion,'' in \emph{Eur. Conf. Comput. Vis.}, 2022, pp. 71--88.

\bibitem{sun_2022_access}
L.~Sun, ``Practical, fast and robust point cloud registration for scene stitching and object localization,'' \emph{IEEE Access}, vol.~10, pp. 3962--3978, 2022.

\bibitem{dai2009rotation}
Y.~Dai, J.~Trumpf, H.~Li, N.~Barnes, and R.~Hartley, ``Rotation averaging with application to camera-rig calibration,'' in \emph{Asian Conf. on Computer Vision}, 2009, pp. 335--346.

\bibitem{inna2010arithmetic}
I.~Sharf, A.~Wolf, and M.~Rubin, ``Arithmetic and geometric solutions for average rigid-body rotation,'' \emph{Mechanism and Machine Theory}, vol.~45, no.~9, pp. 1239 -- 1251, 2010.

\bibitem{lam2007precision}
Q.~M. {Lam} and J.~L. {Crassidis}, ``Precision attitude determination using a multiple model adaptive estimation scheme,'' in \emph{IEEE Aerospace Conference}, 2007, pp. 1--20.

\bibitem{markley2007averaging}
L.~Markley, Y.~Cheng, J.~Crassidis, and Y.~Oshman, ``Averaging quaternions,'' \emph{Journal of Guidance, Control, and Dynamics}, vol.~30, pp. 1193--1196, 07 2007.

\bibitem{humbert1996determination}
M.~Humbert, N.~Gey, J.~Muller, and C.~Esling, ``{Determination of a Mean Orientation from a Cloud of Orientations. Application to Electron Back-Scattering Pattern Measurements},'' \emph{Journal of Applied Crystallography}, vol.~29, no.~6, pp. 662--666, 1996.

\bibitem{morawiec1998note}
A.~Morawiec, ``{A note on mean orientation},'' \emph{Journal of Applied Crystallography}, vol.~31, no.~5, pp. 818--819, 1998.

\bibitem{hartley2013rotation}
R.~Hartley, J.~Trumpf, Y.~Dai, and H.~Li, ``Rotation averaging,'' \emph{International Journal of Computer Vision}, 2013.

\bibitem{aftab2015convergence}
K.~{Aftab} and R.~{Hartley}, ``Convergence of iteratively re-weighted least squares to robust m-estimators,'' in \emph{IEEE Winter Conf. on Applications of Computer Vision}, 2015, pp. 480--487.

\bibitem{lajoie_2019_ral}
P.-Y. Lajoie, S.~Hu, G.~Beltrame, and L.~Carlone, ``Modeling perceptual aliasing in {SLAM} via discrete–continuous graphical models,'' \emph{IEEE Robotics and Automation Letters}, vol.~4, no.~2, pp. 1232--1239, 2019.

\bibitem{tukey}
A.~E. Beaton and J.~W. Tukey, ``The fitting of power series, meaning polynomials, illustrated on band-spectroscopic data,'' \emph{Technometrics}, vol.~16, no.~2, pp. 147--185, 1974.

\bibitem{yang_2020_ral}
H.~Yang, P.~Antonante, V.~Tzoumas, and L.~Carlone, ``Graduated non-convexity for robust spatial perception: From non-minimal solvers to global outlier rejection,'' \emph{IEEE Robotics and Automation Letters}, vol.~5, no.~2, pp. 1127--1134, 2020.

\bibitem{manton_2004_icarcv}
J.~Manton, ``A globally convergent numerical algorithm for computing the centre of mass on compact lie groups,'' in \emph{Int. control, automation, robotics and vision conf.}, vol.~3, 2004, pp. 2211--2216.

\bibitem{markely_2007_jgcd}
F.~L. Markley, Y.~Cheng, J.~L. Crassidis, and Y.~Oshman, ``Averaging quaternions,'' \emph{Journal of Guidance, Control, and Dynamics}, vol.~30, no.~4, pp. 1193--1197, 2007.

\bibitem{sarlette_2009_siam}
A.~Sarlette and R.~Sepulchre, ``Consensus optimization on manifolds,'' \emph{SIAM Journal on Control and Optimization}, vol.~48, no.~1, pp. 56--76, 2009.

\bibitem{robust_single_rotation_averaging}
S.~H. Lee and J.~Civera, ``Robust single rotation averaging,'' \emph{CoRR, abs/2004.00732}, 2020.

\bibitem{weiszfeld1}
E.~Weiszfeld, ``Sur le point pour lequel la somme des distances de n points donn{\'e}s est minimum,'' \emph{Tohoku Mathematical Journal}, vol.~43, pp. 355--386, 1937.

\bibitem{weiszfeld2}
E.~Weiszfeld and F.~Plastria, ``On the point for which the sum of the distances to n given points is minimum,'' \emph{Annals of Operations Research}, vol. 167, no.~1, pp. 7--41, 2009.

\bibitem{yang_2020_nips}
H.~Yang and L.~Carlone, ``One ring to rule them all: Certifiably robust geometric perception with outliers,'' in \emph{Advances in Neural Information Processing Systems}, vol.~33, 2020, pp. 18\,846--18\,859.

\bibitem{shi_2021_icra}
J.~Shi, H.~Yang, and L.~Carlone, ``{ROBIN}: a graph-theoretic approach to reject outliers in robust estimation using invariants,'' in \emph{IEEE Intl. Conf. on Robotics and Automation (ICRA)}, 2021, pp. 13\,820--13\,827.

\bibitem{mankovich_2023_arxiv}
N.~Mankovich and T.~Birdal, ``Chordal averaging on flag manifolds and its applications,'' \emph{CoRR}, vol. abs/2303.13501, 2023.

\bibitem{forster2017onmanifold}
C.~Forster, L.~Carlone, F.~Dellaert, and D.~Scaramuzza, ``On-manifold preintegration for real-time visual--inertial odometry,'' \emph{IEEE Trans. Robot.}, vol.~33, no.~1, pp. 1--21, 2017.

\bibitem{arun}
K.~S. Arun, T.~S. Huang, and S.~D. Blostein, ``Least-squares fitting of two 3-{D} point sets,'' \emph{IEEE Trans. Pattern Anal. Mach. Intell.}, vol.~9, no.~5, pp. 698--700, 1987.

\bibitem{eppstein_2010_ac}
D.~Eppstein, M.~L{\"o}ffler, and D.~Strash, ``Listing all maximal cliques in sparse graphs in near-optimal time,'' in \emph{Algorithms and Computation}, 2010, pp. 403--414.

\bibitem{bunny}
B.~Curless and M.~Levoy, ``A volumetric method for building complex models from range images,'' in \emph{SIGGRAPH}.\hskip 1em plus 0.5em minus 0.4em\relax ACM, 1996, pp. 303--312.

\end{thebibliography}
\end{document}